\documentclass{article}
\usepackage{ijcai21}

\usepackage{times}
\usepackage{soul}
\usepackage{url}
\usepackage[hidelinks]{hyperref}
\usepackage[utf8]{inputenc}
\usepackage[small]{caption}
\usepackage{graphicx}
\usepackage{amsmath}
\usepackage{amsthm}
\usepackage{booktabs}
\usepackage{algorithm}
\usepackage{algorithmic}
\usepackage{multirow}

\title{ Graph Consistency Based Mean-Teaching for Unsupervised Domain Adaptive Person Re-Identification}

\author{
    Xiaobin Liu, Shiliang Zhang
    \affiliations
    Department of Computer Science, School of EECS, Peking University
    \emails
    \{xbliu.vmc, slzhang.jdl\}@pku.edu.cn
}

\begin{document}

\maketitle

\begin{abstract}
%Labeled source data is not always available for unsupervised training on target data due to privacy issue in real-world applications.
%While source models pre-trained on labeled source datasets are always public available. This paper focuses on unsupervised training with multiple source models.
%To fuse information provided by multiple models, we propose a Graph Consistency based Mean-Teaching (GCMT) method for unsupervised domain adaptive person Re-Identification (ReID).
 Recent works show that mean-teaching is an effective framework for unsupervised domain adaptive person re-identification. However, existing methods perform contrastive learning on selected samples between teacher and student networks, which is sensitive to noises in pseudo labels and neglects the relationship among most samples.
 Moreover, these methods are not effective in cooperation of different teacher networks.
 To handle these issues, this paper proposes a Graph Consistency based Mean-Teaching (GCMT) method with constructing the Graph Consistency Constraint (GCC) between teacher and student networks. 
  Specifically, given unlabeled training images, we apply teacher networks to extract corresponding features and further construct a teacher graph for each teacher network to describe the similarity relationships among training images. To boost the representation learning, different teacher graphs are fused to provide the supervise signal for optimizing student networks.
%The advantage of using the graph consistency is that it can model more relationship among different samples.  
%  initializes a pair of student and teacher networks for each source models. Features of unlabeled samples extracted by each teacher network are formed into a KNN graph to describe the relationship among different samples. Graphs constructed by different teacher networks are fused as supervise signal for optimizing student networks.
%   the teacher network to extract features of unlabeled samples and further construct a KNN graph  The constructed graph is used as the supervise signal for  
 GCMT  fuses similarity relationships predicted by different teacher networks as supervision and effectively optimizes student networks with more sample relationships involved.
 Experiments on three datasets, \textit{i.e.}, \textit{Market-1501}, \textit{DukeMTMC-reID}, and \textit{MSMT17}, show that proposed GCMT outperforms state-of-the-art methods by clear margin.
 Specially, GCMT even outperforms the previous method that uses a deeper backbone.
% Specially, with training starting from the model pre-trained on \textit{ImageNet}, GCMT still outperforms most recent methods.
% with source models on other person datasets. 
 Experimental results also show that GCMT can effectively boost the performance with multiple teacher and student networks.
 Our code is available at https://github.com/liu-xb/GCMT .

%The advantage of using the graph consistency  is that it 
% To handle these issues, this paper proposes a label-free Graph Consistency based Mean-Teaching (GCMT) method for contrastive learning in feature space. Specifically, we construct a complete graph whose nodes are features of unlabeled samples, and relationship among samples is represented as weights of edges between nodes, \textit{i.e.}, structure of the graph. Then, GCMT encourages the student to learn the graph structure from the mean teacher by pulling the adjacency matrix of student graph close to the adjacency matrix of teacher graph. GCMT takes into consideration the global sample relationship without using noisy pseudo labels, thus exhibits effective contrastive learning. 

%To handle these issues, we propose a label-free Graph-Consistency-based Mean-Teaching (GCMT) for unsupervised person re-identification with constructing an graph consistency constraints between teacher and student networks.
%Given the batch images, we apply the teacher network to extract the corresponding features and further construct a undirected graph to describe the similarity among different samples. To boost the representation learning,  the constructed graph is treated as the supervise signal  for optimizing the student networks.
%The advantage of using the graph consistency  is that it can model the more relationship among different samples.
\end{abstract}

\section{Introduction}
\label{sec:intro}
Person Re-Identification (ReID) aims to match a query person image in a gallery set~\cite{xuan2021intra,li2020joint,li2019pose,wei2018vp}. Supervised ReID has been widely studied from different aspects, such as parts feature extraction~\cite{liu2018ram,jianing}, metric learning~\cite{zhou2017point,liu2019group}, attention learning~\cite{Zhang_2020_CVPR,teng2018scan}, \textit{etc.}
However, supervised models suffer from the expensive data annotation and substantial performance drop when applied on different target domains. To address these issues, recent works focus on unsupervised domain adaptive person ReID and exhibit promising performance. More details of recent works are summarized in Sec.~\ref{sec:related}.

\begin{figure}[t]
\begin{center}
\includegraphics[width=1\linewidth]{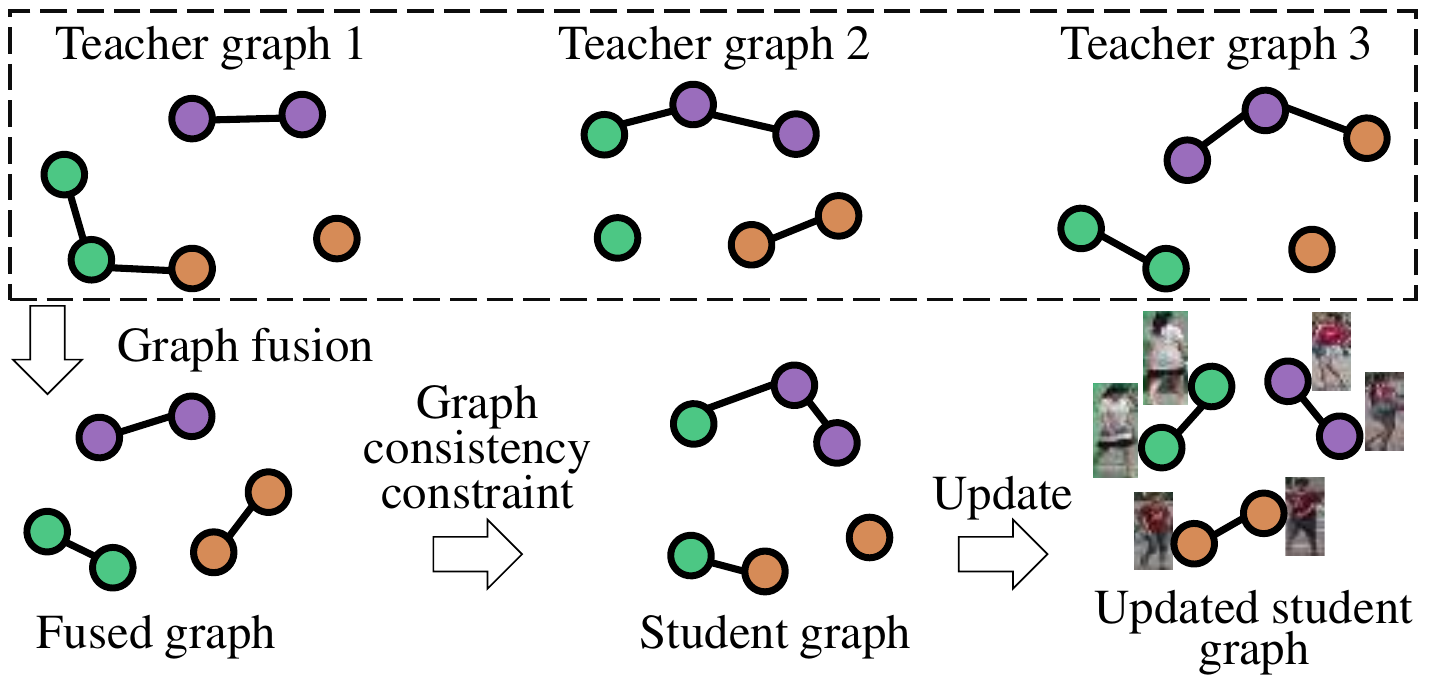}
\end{center}
\vspace{-5mm}
\caption{Illustration of Graph Consistency Constraint (GCC) in proposed GCMT method between three teacher graphs and one student graph. Relationships of samples in the student graph are updated to approach relationships in the graph fused by multiple teacher graphs. Dots in different colors denote different identities.}
\label{fig:intro}
\vspace{-3mm}
\end{figure}

Despite the significant success, %unsupervised domain adaptive person ReID is still a challenging task and
 there still remain several open issues unexplored. 
 Firstly, previous works commonly perform training on both labeled source data and unlabeled target data~\cite{dimglo}. In real-world applications, however, it is always impractical to access  the labeled source data due to privacy issue. Moreover, labeled datasets are also more expensive to store and process than pre-trained models. This hinders the training of these methods.
 Secondly, person ReID model optimized by mean-teaching framework requires an effective student network training strategy that should be robust to noisy labels and also effective in feature learning. Existing methods use soft triplet loss on selected samples based on pseudo labels~\cite{mebnet}. However, it is sensitive to noisy pseudo labels and ineffective in feature learning with few selected sample relationships involved.
 Thirdly, models pre-trained on different labeled datasets are public available and using them in training has potential to boost the performance on target datasets. However, current methods with soft triplet loss are not effective in collaborative learning with multiple models.

This paper is motivated to study an effective training method for unsupervised domain adaptive person ReID. To release the dependence on labeled source data, only unlabeled target data is used in unsupervised training.
 To enhance the effectiveness in feature learning of student networks and cooperation of teacher networks, we propose a Graph Consistency based Mean-Teaching (GCMT) method with a Graph Consistency Constraint (GCC) between teacher and student networks as shown in Fig.~\ref{fig:intro}.
 GCMT merges sample relationships predicted by different teacher networks as supervision and optimizes more relationships among different samples in student networks, thus provides effective supervision and optimization for feature learning on student networks. 
%Our method is insensitive to the pre-trained model that training starts from.
 
As illustrated in Fig.~\ref{fig:intro}, the proposed Graph Consistency Constraint (GCC) in GCMT method is performed between teacher graphs and student graphs.
 Specifically, given public pre-trained models, GCMT initializes a  student network and its temporal mean teacher network from each model. 
 Features extracted by a teacher network are formed into a teacher graph with using features as nodes. Connections between nodes in teacher graphs are determined by KNN strategy, \textit{i.e.}, a node is only connected with its KNN nodes. Weights of connections are computed as feature similarities.
 Features extracted by a student network are formed into a student graph with weights of connections also computed as feature similarities.
 Teacher graphs are fused to guide the connection weight leaning in student graphs by GCC via encouraging the connection weights in student networks to approach the weights in the fused graph. Thus, GCC guides student graphs to describe the same sample relationship with the fused teacher graph.
 Compared with previous methods, proposed GCC avoids the dependency on pseudo labels.
 It involves relationships among more samples in feature learning and also can effectively fuse sample relationships predicted by multiple teacher networks, hence is more effective in feature learning and also robust against noises in pseudo labels.
To the best of our knowledge, it is the first work to use GCC in mean-teaching framework for unsupervised training.

%Graph consistency constraint is flexible to extend when multiple models pre-trained on different datasets are given. Specifically, a teacher network and a student network are built for each pre-trained model. Weights in different teacher graphs are averaged to supervise the updating of student graphs. Thus, GCMT is easy to perform model ensemble, \textit{i.e.}, starting training from multiple models pre-trained on different datasets to further boost the performance. 

We test our method on three large-scale datasets, \textit{i.e.}, \textit{Market-1501}, \textit{DukeMTMC-reID}~, and \textit{MSMT17}. Comparison with recent works shows that our method outperforms state-of-the-art methods by a clear margin. For instance, training from model pre-trained on \textit{ImageNet}, GCMT achieves mAP accuracy of 63.6\% on \textit{DukeMTMC-reID}, outperforming recent UTAL~\cite{li2019unsupervised-pami} by 19\%. %It is worth noting that our GCMT also outperforms recent MMT~\cite{mmt} by 1.8\% which uses source model on \textit{Market-1501}. 
 By using model pre-trained on \textit{DukeMTMC-reID}, our GCMT method achieves mAP accuracy of 77.1\% on \textit{Market-1501}, significantly outperforming recent MEB-Net~\cite{mebnet} and MMT~\cite{mmt} by 4.9\% and 5.9\%, respectively.
 Experiments also show that GCMT boosts the performance with multiple pre-trained models, \textit{e.g.}, the mAP accuracy is boosted to 79.7\% on \textit{Market-1501} with models pre-trained on \textit{MSMT17}, \textit{CUHK03}, and \textit{DukeMTMC-reID}.

%In summary, this work propose GCMT method for unsupervised domain adaptive person ReID. GCMT constructs directed graphs for teacher and student, respectively, and encourages the student to learn the same graph structure as teacher. Compared with existing works, GCMT is effective in knowledge distillation without using pseudo labels. This guarantees the superior performance of our method in comparisons with state-of-the-art works.

\section{Related Work}
\label{sec:related}

%This work is related with unsupervised domain adaptation and unsupervised domain adaptive person ReID. This section briefly reviews recent works on these tasks.

%\subsection{Unsupervised Domain Adaptation} 
%Works on Unsupervised Domain Adaptation (UDA) mainly target to tackle the domain gap issue~\cite{kumagai2019unsupervised,tzeng2017adversarial}. 
%For example, Kumagai \textit{et al.}~\cite{kumagai2019unsupervised} propose to transform the feature distribution of labeled domain to approach the feature distribution of unlabeled domain. 
%Tzeng \textit{et al.}~\cite{tzeng2017adversarial} use adversarial loss to encourage the feature distribution of unlabeled dataset to approach the feature distribution of labeled dataset. UDA is similar with domain adaptive person ReID. However, UDA holds an assumption that the unlabeled domain has same categories with the labeled domain. While, in unsupervised domain adaptive person ReID, labeled and unlabeled domains usually have totally different identities. Thus, these methods are not suitable for ReID task.

%\subsection{Unsupervised Domain Adaptive Person ReID}

%\textbf{Domain-invariant feature learning.}
\paragraph{Domain-invariant feature learning.}
 Some works design GAN based models to transfer labeled images to target domains~\cite{Liu_2019_CVPR}. 
Several works map labeled and unlabeled images to a shared feature space to bridge the domain gap~\cite{dimglo}. 
These methods require labeled source data for training, which is hardly available due to privacy issue in real-world applications. Compared with them, this work only uses pre-trained models instead of labeled data, thus is more suitable for real-world applications.

%\textbf{Self-supervised learning.} 
\paragraph{Self-supervised learning.} 
Some works locally predict pseudo labels for unlabeled samples~\cite{zhong2019invariance,yu2019unsupervised}.
To acquire reliable pseudo labels, some researchers try to refine unsupervised clustering~\cite{ding2019towards,lin2019bottom}. 
For unsupervised optimization, current works use pseudo labels as supervision and adopt triplet loss~\cite{ssg,Zhang_2019_ICCV} or contrastive loss~\cite{dimglo,zhong2019invariance} for training. 
 However, these methods fail to consider the noise in pseudo labels, which substantially hinders the model training. Compared with them, proposed model adopts fused graph of temporal mean teacher networks to supervise feature learning, alleviating the effect of noisy labels and improves the performance. 
 Ge \textit{et al.}~\cite{ge2020self} propose a self-paced learning method to boost the ReID performance. However, they have different motivations and propose different algorithms compared with this paper. They aim to make full use of labeled data by caching features in a hybrid memory bank. While, this paper aims to study an effective mean-teaching method on unlabeled datasets without labeled data by proposing a graph consistency constraint. Experiments show our method achieves better performance under the same setting.

\paragraph{Teacher-student training.} 
Teacher-student framework is widely studied in semi-supervised learning~\cite{tarvainen2017mean,han2018co}.%, \textit{e.g.}, mean teaching method~\cite{tarvainen2017mean} uses temporal average model as teacher to provide supervision for training.
 However, these methods hold the assumption that labeled and unlabeled images are of same categories, making them not suitable for person ReID task.
 Recently, some works adopt mean-teaching method in unsupervised domain adaptive person ReID. MMT~\cite{mmt} uses two pairs of teacher and student networks for pseudo label refinery. MEB-Net~\cite{mebnet} further uses three pairs of teacher and student networks of three different structures for training. 
 MMT and MEB-Net use triplet loss in model training and soft triplet loss between teacher and student networks, which are sensitive to noisy label and ineffective in feature learning. Moreover, training with multiple teacher networks is still ineffective and the improvements by using three teacher networks is limited in~\cite{mebnet}. 
 Compared with them, proposed GCMT is effective in both student network training and teacher network cooperation. Experimental results show that GCMT achieves better performance compared with these methods.

%\textbf{Additional cues.} Some works adopt additional cues to boost performance. For example, SSG~\cite{ssg} and PAST~\cite{Zhang_2019_ICCV} use local features, CAL~\cite{Qi_2019_ICCV} and UTAL~\cite{li2019unsupervised-pami} use temporal information. Experiments show that our method outperforms these methods by a clear margin without additional cues.

%-------------------------------------------------------------------------

\section{Problem Formulation}
\label{sec:formulation}
Given an unlabeled dataset $\mathcal{D}$, unsupervised domain adaptive person ReID aims to learn the ReID model on $\mathcal{D}$ without annotation. $\mathcal{D}$ can be denoted as $\{ x_i | i = 1...N \}$. $x_i$ and $N$ denote the $i$-th image and the number of images in $\mathcal{D}$, respectively. In this paper, training starts from public models pre-trained on different datasets, while labeled source data is not used. This setting is more practical in real-world scenario compared with several previous works as discussed in Sec.~\ref{sec:intro}. Public pre-trained models are denoted as $\{ M^1, M^2, ..., M^m\}$ where $m$ denotes the number of models.

\begin{figure}[t]
\begin{center}
\includegraphics[width=0.99\linewidth]{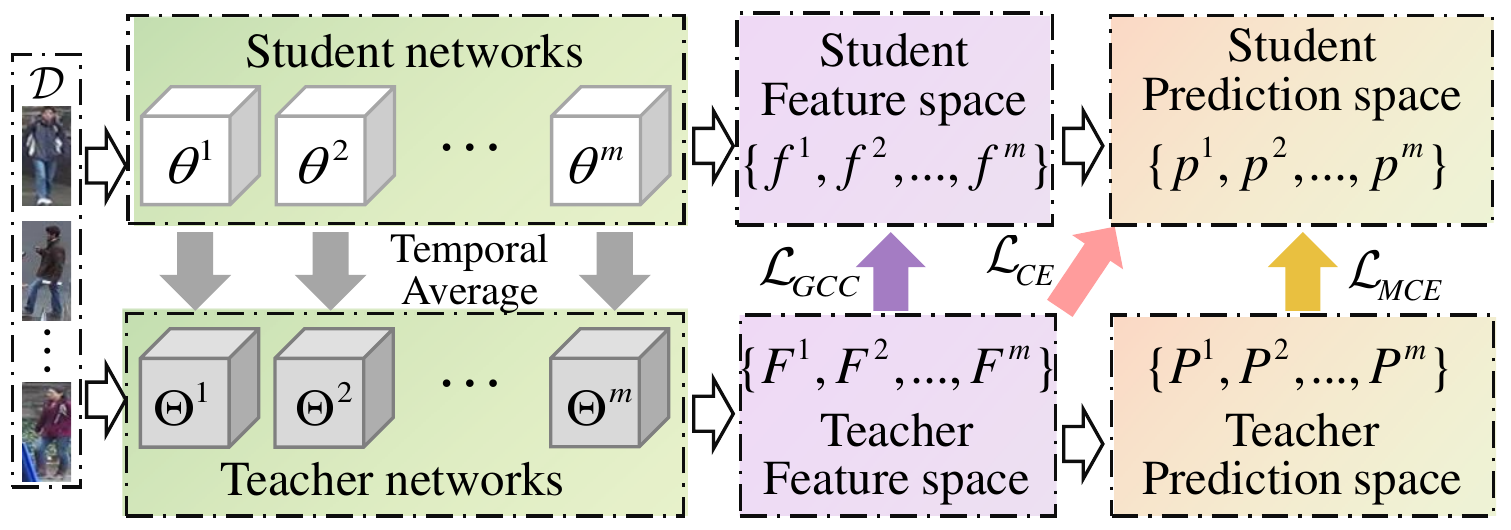}
\end{center}
\vspace{-4mm}
\caption{Framework of proposed GCMT method. $m$ pairs of student and teacher networks are initialized by $m$ public pre-trained models. Teacher network $\Theta^j$ is updated by temporal average strategy with corresponding student network $\theta^j$. Features extracted by student networks are supervised by features extracted by teacher networks via graph consistency constraint $\mathcal{L}_{GCC}$. Class predictions by student networks are supervised by pseudo labels via cross entropy loss $\mathcal{L}_{CE}$ and predictions by teacher networks via mutual cross entropy loss $\mathcal{L}_{MCE}$. Pseudo labels are generated by off-line clustering on features extracted by teacher networks.}
\label{fig:framework}
\vspace{-4mm}
\end{figure}

 We adopt mean-teaching framework for unsupervised training on $\mathcal{D}$ as shown in Fig.~\ref{fig:framework}. $m$ student networks are initialized by $m$ public pre-trained model, respectively.
 The parameters of student networks are denoted as $\{\theta^1, \theta^2, ..., \theta^m \}$ with $\theta^j$ denoting the parameters of the $j$-th student network. We use temporal average network as the teacher network for each student network as in~\cite{tarvainen2017mean}. Parameters in teacher networks are denoted as $\{ \Theta^1, \Theta^2,..., \Theta^m\}$, respectively. In the beginning of training, $\theta^j$ and $\Theta^j$ are both initialized by $M^j$. In each iteration of the training, $\theta^j$ is updated by objective functions and $\Theta^j$ is then updated by temporal average strategy with updated $\theta^j$ as:
\begin{eqnarray}
\label{eq:mmt}
\Theta^j \leftarrow 0.999 \Theta^j + 0.001\theta^j .
\end{eqnarray}
Coefficients in Eqn.~\eqref{eq:mmt} are empirical values following~\cite{mmt,mebnet}.

 For unlabeled images $x_i$, the extracted feature after L2-normalization and class prediction after softmax normalization by the $j$-th student network $\theta^j$ are denoted as $f_i^j$ and $p_i^j$, respectively. 
 And the extracted feature and prediction by the $j$-th teacher network $\Theta^j$ for $x_i$ are denoted as $F_i^j$ and $P_i^j$, respectively.
 Note that we use the superscript and subscript to denote the index of networks and images, respectively.

 The target of training on $\mathcal{D}$ is to make extracted features discriminative for person ReID task. In this paper, we propose a Graph Consistency based Mean Teaching (GCMT) method to achieve this goal via constraints in two spaces: 1) class prediction space, and 2) feature space. The framework of proposed method is illustrated in Fig.~\ref{fig:framework}.

 In class prediction space, we first generate pseudo ID labels on $\mathcal{D}$ by unsupervised clustering on averaged features extracted by teacher networks before each epoch. Based on pseudo labels, cross entropy loss $\mathcal{L}_{CE}$ is computed for pseudo ID classification on each student network.
 We use a new fully connected layer for label prediction after each clustering step following~\cite{mmt,mebnet}. After clustering features into $C$ clusters, a mean feature for each cluster is computed by averaging all features in this cluster. This results in $C$ mean features corresponding to $C$ clusters.
Then, a new fully connected layer with $C$-way outputs is used for label prediction and parameters are initialized with corresponding mean features, \textit{i.e.}, the parameters for the $c$-th class is initialized by the mean feature of the $c$-th cluster.

  To eliminate the negative effects of noises in pseudo label inside each training batch, mutual cross entropy loss $\mathcal{L}_{MCE}$ is used to encourage student networks to predict same class probabilities with teacher networks for training samples. Specifically, the probabilities of $x_i$ by different teacher networks are averaged as the training target for each student network, which is denoted as $\hat{P}_i=\sum_j^mP_i^j$. The objective function of $\mathcal{L}_{MCE}$ with $m$ teacher networks is formulated as:
\begin{eqnarray}
\mathcal{L}_{MCE} = -\frac{1}{N}\sum_{i=1}^N \sum_{c=1}^C \hat{P}_i(c)\sum_{j=1}^m \log p_i^j(c),
\end{eqnarray}
where $C$ denotes the number of identities generated by clustering. $\hat{P}_i(c)$ and $p_i^j(c)$ denote the probability of $x_i$ being classified to $c$-th class in $\hat{P}_i$ and $p^j_i$, respectively.
 The objective function in label prediction space $\mathcal{L}_{LP}$ is the combination of $\mathcal{L}_{CE}$ and $\mathcal{L}_{MCE}$ as:
\begin{eqnarray}
\mathcal{L}_{LP}=\mathcal{L}_{CE} + \mathcal{L}_{MCE}.
\end{eqnarray}

 Classification probabilities describe the relationship between samples and classes. Thus, $\mathcal{L}_{LP}$ focuses on classifying training samples to corresponding pseudo labels. For ReID task, however, we also need to optimize the model in feature space, as identities in testing set are different with training set and ReID is performed as retrieval by extracted features. Thus, we propose Graph Consistency Constraint (GCC) between teacher and student networks in feature space.
 
 As shown in Fig.~\ref{fig:framework}, GCC is proposed to train student networks in feature space. 
 GCC computes sample similarity relationship for each network with corresponding extracted features, respectively. Sample similarity relationships predicted by different teacher networks are fused as the target relationship. Student networks are encouraged by GCC to output features that have the same sample similarity relationship as the target relationship as shown in Fig.~\ref{fig:intro}. The objective function of GCC is denoted as $\mathcal{L}_{GCC}$ and details of the computation on $\mathcal{L}_{GCC}$ will be given in Sec.~\ref{sec:gc}.

$\mathcal{L}_{LP}$ is performed in class prediction space and focuses on relationship between samples and pseudo IDs. While $\mathcal{L}_{GCC}$ is performed in feature space and focuses on similarity relationship among samples.
Therefore, $\mathcal{L}_{LP}$ and $\mathcal{L}_{GCC}$ are complementary to each other. We hence denote the final objective function as the combination of $\mathcal{L}_{LP}$ and $\mathcal{L}_{GCC}$:%, \textit{i.e.},
\begin{eqnarray}
\label{eqn:l}
\mathcal{L} = \mathcal{L}_{LP} + \lambda_{GCC}\mathcal{L}_{GCC},
\end{eqnarray}
where $\lambda_{GCC}$ is the loss weight for $\mathcal{L}_{GCC}$.

\section{Graph Consistency Constraint}
\label{sec:gc}
In feature learning, we aim to fuse sample similarity relationships computed by different teacher networks to provide training supervision for student networks.
 Features extracted by different teacher networks are of different distributions. Thus, directly averaging similarities computed by different teacher networks is not reasonable for relationship fusion. Moreover, when only relationship between several selected samples are involved, \textit{e.g.}, soft triplet loss with hard sample mining in~\cite{mebnet}, noise in pseudo labels will mislead both the sample selection and optimization.

 To chase a reasonable relationship fusion and an effective student network optimization, we propose the Graph Consistency Constraint (GCC) that describes sample similarity relationship as weights of edges in graphs. Then, GCC performs relationship fusion and network optimization based on graphs.
 By fusing relationship predicted by different teacher networks and involving more samples in training student networks, GCC eliminates the effect of noises in pseudo labels and provides effective update for student networks.
 
\subsection{Teacher Graph Construction} 
Similarity relationship predicted by each teacher network is described as a graph. Specifically, a $K$-nearest neighbour graph $G(F^j, W^j)$ is constructed for features extracted by the $j$-th teacher network $\Theta^j$.  $F^j=\{F_i^j|i=1...N \}$ and $W^j=\{ W_{i,k}^j|i,k=1...N \}$ denote the set of features of nodes and the set of weights of edges between nodes, respectively. Weights of edges describe the distance relationships among different features in the graph, which are computed based on $K$-NN relationship.
 Specifically, $F_k^j$ is connected to $F_i^j$ if $F_k^j$ is among the $K$-nearest neighbours of $F_i^j$. The weight of edge from $F_i^j$ to $F_k^j$, \textit{i.e.}, $W^j_{i,k}$, is computed as $W^j_{i,k}= (F_i^j)^T F_k^j$. For pairs of nodes that are not connected, corresponding weights of edges are assigned as 0.
 
 Graphs constructed by different teacher networks are fused to provide training target for student networks training. Weights in each graph are first normalized by softmax and then averaged to obtain the fused weights as follows:
\begin{eqnarray}
\hat{W}_{i,k}=\frac{1}{m}\sum_{j=1}^m \frac{\exp(W_{i,k}^j)}{\sum_{h=1}^{N,h\not=i}\exp(W_{i,h}^j)}.
\end{eqnarray} 
 The set of fused weights $\hat{W}_{i,k}$ is denoted as $\hat{W}$.
  
\subsection{Graph Consistency Constraint Computation}
Student networks are supervised by GCC to extract features that have the same similarity relationships between  different samples described by $\hat{W}$.
 The computation of GCC should satisfy three conditions.
 1): GCC should simultaneously update every relationship between different training samples for student networks, making optimization effective and precise.
 2): GCC should be aware of the density of neighbours. Some samples have dense neighbours whose $K$-nearest neighbours should reach rather high similarities with them. While some other samples have sparse neighbours, and restriction on their neighbours can be relaxed. Thus, GCC should optimize relative similarity between samples, instead of absolute similarity.
 3): GCC should be aware of the hardness of $K$-nearest neighbours of each sample, \textit{i.e.}, gradients of far neighbours should be larger than gradients of near neighbours.

For computing GCC, student graph $G(f^j, w^j)$ is constructed for features extracted by the $j$-th student network $\theta^j$, where $f^j=\{f^j_i|i=1...N \}$ and $w^j=\{w^j_{i,k}| i,k=1...N \}$ denote the set of features of nodes and the set of weights of edges between nodes, respectively.
 Weights of edges from $f_i^j$ to $f_k^j$ in $G(f^j, w^j)$ is computed as:
 \begin{eqnarray}
 \label{eq:w} 
 w^j_{i,k} = \frac{\exp((f^j_i)^T f^j_k / \beta )}{\sum_{h=1}^{N,h\not=i} \exp((f^j_i)^T f^j_h / \beta )} ,
 \end{eqnarray}
 where $\beta$ is a hyper-parameter to adjust the scale of distribution. GCC is computed to encourage $\{w^j|j=1...m\}$ to approach $\hat{W}$ via cross entropy which is formulated as: 
\begin{eqnarray}
\label{eq:gc}
\mathcal{L}_{GCC}= -\frac{1}{NK}\sum_{i=1}^N \sum_{k=1}^{N,k\not=i} \hat{W}_{i,k} \sum_{j=1}^m \log w^j_{i,k}.
\end{eqnarray}

Proposed objective function in Eqn.~\eqref{eq:gc} satisfies aforementioned three conditions. 
 1): Every edge in student graphs is updated by $\mathcal{L}_{GCC}$ based on $\hat{W}$. % $w^j_{i,k}$ is enlarged if $\hat{W}_{i,k} > 0$, and reduced if $\hat{W}_{i,k} = 0$ because $w^j_{i,k}$ is computed by softmax normalization in Eqn.~\eqref{eq:w}.
 2): Due to the softmax normalization in Eqn.~\eqref{eq:w}, $\mathcal{L}_{GCC}$ normalizes loss of samples with different densities of neighbours to the same scale. Thus, $\mathcal{L}_{GCC}$ is aware of the density of neighbours.
 3): The gradient of $\mathcal{L}_{GCC}$ relative to $w^j_{i,k}$ when $\hat{W}_{i,k} \neq 0$ is computed as:
$
\frac{\partial \mathcal{L}_{GCC}}{\partial w^j_{i,k}} = -\frac{\hat{W}_{i,k}}{N w^i_{i,k}}.
$ %\end{eqnarray}
 The scale of gradient increases with $w^j_{i,k}$ decreasing. This indicates that $\mathcal{L}_{GCC}$ pays more attention on $K$-nearest neighbours with small similarities for each sample. Thus, $\mathcal{L}_{GCC}$ is also aware of hardness.

\subsection{Discussion}
 GCMT fuses sample similarity  relationship computed by different teacher networks as supervise signal  instead of pseudo labels. As networks are being more discriminative with training, similarity relationship computed in each iteration has potential to be more precise than pseudo labels computed off-line before each epoch.
 Moreover, $\mathcal{L}_{GCC}$ avoids sample mining algorithm and automatically focuses on hard neighbours.

 Compared with soft triplet loss used in~\cite{mebnet} and~\cite{mmt} that  trains student networks based on pseudo labels, $\mathcal{L}_{GCC}$ optimizes distance relationship in student graphs. This considers more relationship among samples and avoids the negative effect of noises in pseudo labels. Specially, soft triplet loss can be regarded as a special case of GCC that uses sub-graphs with selected samples. Advantages of our method will be shown in Sec.~\ref{sec:compared}.

Compared with MEB-Net~\cite{mebnet} that fuses relationship predicted by different teacher networks via soft triplet loss based on pseudo labels, GCC performs relationship fusion based on graphs. This avoids effect of noises in pseudo label and involves more relationship in training. Advantages of our method in multiple teacher networks cooperation will be shown in Sec.~\ref{sec:ma}. %Moreover, graph fusion is efficient in computation and thus flexible to be extended with multiple source models. 

%comparison
\begin{table*}[ht]
%\vspace{-5mm}
\begin{center}
\footnotesize
\resizebox{1.\linewidth}{!}{
\begin{tabular}{l|c|c|c|c|c|c||c|c|c|c|c|c}
\hline
%\multirow{2}{*}{Method} & \multirow{2}{*}{Reference} & \multicolumn{4}{c||}{Market-1501} & \multicolumn{4}{c}{DukeMTMC-reID}\\
\multirow{2}{*}{Method} & %\multirow{2}{*}{B.}&
 \multicolumn{6}{c||}{Market} & \multicolumn{6}{c}{Duke}\\

\cline{2-13}
 &S. M. & S. D. &mAP & Rank1 & Rank5 & Rank10 &S. M. & S. D. & mAP & Rank1 & Rank5 & Rank10 \\
%\hline
%Supervised %& Baseline
%& 0.728 & 0.899& 0.969 & 0.981 
%&0.616 & 0.770 & 0.895 & 0.921 \\

\hline
Supervised baseline %& R.50 
& $M^I$ & \textit{Market} & 0.811 & 0.930 & 0.974 & 0.985
& $M^I$ & \textit{Duke} & 0.704 & 0.848 & 0.920 & 0.943\\
\hline

%LOMO~\cite{lomo} %& CVPR
%&0.080 &0.272&0.416&0.491
%&0.048 &0.123 &0.213&0.266\\

%BOW~\cite{market} %& ICCV
%&0.148 &0.358&0.524&0.603
%&0.083 &0.171&0.288&0.349\\

DBC~\cite{ding2019towards} %& \multirow{15}{*}{R.50}%& BMVC
&$M^I$ & None&0.413&0.692&0.830&0.878
&$M^I$ & None&0.300&0.515&0.646&0.701\\

%BUC~\cite{lin2019bottom} %& %& AAAI
%&$M^I$ & None &0.383&0.662&0.796&0.845
%&$M^I$ & None &0.275&0.474&0.626&0.684\\

GLO~\cite{dimglo} %& %& ACMMM
&$M^I$ & None &{0.457}&{0.774}&{0.880}&{0.901}
&$M^I$ & None &{0.364}&{0.605}&{0.722}&{0.757}\\

UTAL~\cite{li2019unsupervised-pami}%&  %& PAMI
&$M^I$ & None&0.462&0.692&-&-
&$M^I$ & None&0.446&0.623&-&-\\

SpCL~\cite{ge2020self}
&$M^I$ & None& 0.731 &0.881&0.951&0.970
&-&-&-&-&-&-\\

GCMT %\multirow{2}{*}{R.50} %& This paper
&$M^I$ & None&\textbf{0.739} &\textbf{0.897}&\textbf{0.965} & \textbf{0.976}
&$M^I$ & None& \textbf{0.636} & \textbf{0.782} &\textbf{0.886} & \textbf{0.913}\\

\hline

%TAUDL$^{\dagger}$~\cite{li2018unsupervised}% & ECCV
%&0.412&0.637&-&-
%&0.435&0.617&-&-\\

%MAR~\cite{yu2019unsupervised}%&CVPR
%&0.400&0.677&0.819&-
%&0.480&0.671&0.798&-\\
%PAUL~\cite{yang2019patch}%& CVPR
%&0.401&0.685&0.824&0.874
%&0.532&0.720&0.827&0.860\\

%CASCL~\cite{wu2019unsupervised}%& %&ICCV
%&$M^{MS}$ & None &0.355&0.654&0.806&0.862
%&$M^{MS}$ & None &0.378&0.593&0.732&0.778\\

%HHL~\cite{zhong2018generalizing} %& ECCV
%&0.314& 0.622&0.788&0.840
%&0.272&0.469&0.610&0.667\\

%ECN~\cite{zhong2019invariance} %& %& CVPR
%&$M^I$ & \textit{Duke} &0.430&0.751&0.876&0.916
%&$M^I$& \textit{Market}&0.404&0.633&0.758&0.804\\

%PAUL~\cite{yang2019patch}%& CVPR
%&0.368& 0.667&-&-
%&0.357&0.561&-&-\\

ATNet~\cite{Liu_2019_CVPR} %& %&CVPR
&$M^I$ & \textit{Duke} &0.256&0.557&0.732&0.794
&$M^I$& \textit{Market} &0.249&0.451&0.595&0.642\\

%CR\_GAN\cite{Chen_2019_ICCV}%&  %& ICCV
%&$M^I$ & \textit{Duke} &0.540& 0.777&0.897&0.927
%&$M^I$& \textit{Market} &0.486&0.689&0.802&0.847\\

%CASCL~\cite{wu2019unsupervised}%&  %&ICCV
%&0.356&0.647&0.802&0.856
%&0.305&0.515&0.667&0.717\\

%DA\_2S~\cite{huang2019sbsgan}%&ICCV
%&0.273&0.585&-&-
%&0.308&0.535&-&-\\
%CAL$^{*}$~\cite{Qi_2019_ICCV}%&ICCV
%&0.345&0.643&-&-
%&0.367&0.554&-&-\\
%CAL~\cite{Qi_2019_ICCV}%& %&ICCV
%&$M^I$ & \textit{Duke} &0.496&0.737&-&-
%&$M^I$ & \textit{Market} &0.456&0.640&-&-\\

PDA-Net~\cite{Li_2019_ICCV}%&  %&ICCV
&$M^I$ & \textit{Duke} & 0.476&0.752&0.863&0.902
&$M^I$ & \textit{Market} &0.451&0.632&0.770&0.825\\

%PAST~\cite{Zhang_2019_ICCV}%&  %&ICCV
%&$M^I$ & \textit{Duke} & 0.546&0.784&-&-
%&$M^I$ & \textit{Market} &0.543&0.724&-&-\\

DIM+GLO~\cite{dimglo} %& %& ACMMM
&$M^I$ & \textit{Duke} &{0.651}&{0.883}&{0.947}&{0.963}
&$M^I$ & \textit{Market}&{0.583}&{0.762}&{0.857}&{0.885}\\

%UDA~\cite{adaptive-reid}%&  %& PR
%&$M^I$ & \textit{Duke} & 0.537&0.758&0.895&0.932
%&$M^I$ & \textit{Market}&0.490&0.684&0.801&0.835\\

%GPP~\cite{zhong2019learning} %& TPAMI
%&0.638& 0.841&0.928&0.954
%&0.544&0.740&0.837&0.874\\
%SpCL~\cite{ge2020self}
%&$M^I$ & \textit{MSMT} & 0.775 & 0.897 & 0.961 & 0.976
%& - & - & - & - & - & - \\

\hline

SSG~\cite{ssg}%& % & ICCV
&$M^D$ & None&0.583&0.800&0.900&0.924
&$M^M$ & None&0.534&0.730&0.806&0.832\\

MEB-Net$^{\S}$
~\cite{mebnet} %& D.121 %& ECCV
&$M^D$ & None&0.760 & 0.899 & 0.960 & 0.975   
&$M^M$ & None& 0.661 & 0.796 & 0.883 & 0.922 \\

MEB-Net~\cite{mebnet} %& \multirow{3}{*}{R.50}%& ECCV
&$M^D$ & None &0.722 & - & -&-
& - & - & - & - &- & - \\

%Co-teaching~\cite{han2018co} %&
%&$M^D$ & None & 0.651 & 0.825 & 0.918 & 0.934
%&$M^M$ & None & 0.557 & 0.719 & 0.835 & 0.881\\ 

MMT~\cite{mmt} %&  %& ICLR
&$M^D$ & None&0.712&0.877&0.949&0.969
&$M^M$ & None&0.631&0.768&0.880&0.922\\
%MMT-700~\cite{mmt} & ICLR'20
%&0.690&0.868&0.946&0.969
%&0.651&0.780&0.888&0.925\\
%MMT$^{\S}$~\cite{mmt} & ICLR
%&0.765&0.909&0.964&0.979
%&0.657&0.793&0.891&0.924\\
%MMT-700$^{\S}$~\cite{mmt} & ICLR'20
%&0.745&0.911&0.965&0.982
%&0.687&0.818&0.912&0.934\\
\hline

GCMT %& %& This paper
&$M^D$ & None&\textbf{0.771}&\textbf{0.906}&\textbf{0.963}&\textbf{0.977}
&$M^M$ & None&\textbf{0.678}&\textbf{0.811}&\textbf{0.911}&\textbf{0.939}\\

\hline

\end{tabular}}
\vspace{-3mm}
\caption{Comparison with state-of-the-art methods on \textit{Market} and \textit{Duke}. ``S. M.'' and ``S. D.'' denote source model and labeled source data, respectively. $^{\S}$ denotes DenseNet121 is used as backbone.} %``$^{\dagger}$'' denotes  temporal information is used. ``$^{\ddagger}$'' denotes multi-scale features are used. }
\label{tab:compared-others}
\end{center}
\vspace{-5mm}
\end{table*}

%msmt
\begin{table}[t]
%\vspace{-5mm}
\begin{center}
\small
\resizebox{1.\linewidth}{!}{
\begin{tabular}{l|c|c|c|c}
\hline
Method &  S. M. & S. D. &  mAP & Rank1\\
\hline 
%Supervised baseline  & $M^I$ &\textit{MSMT} & 0.463 & 0.736 \\
SpCL~\cite{ge2020self}& $M^I$ &None &0.191 & 0.423\\

GCMT & $M^I$ & None & 0.237 & 0.543 \\
GCMT & $M^C$ & None & 0.263 & 0.573 \\
\hline
%\hline
ECN~\cite{zhong2019invariance}  &$M^I$&\textit{Market} & 0.085 & 0.253\\

DIM+GLO~\cite{dimglo} & $M^I$ & \textit{Market} & 0.207 & 0.497 \\
%
%SSG~\cite{ssg} & $M^M$ & None & 0.132 & 0.316\\
%
MMT~\cite{mmt} & $M^M$ & None & 0.229 & 0.492\\
GCMT & $M^M$ & None & \textbf{0.249} & \textbf{0.548 }\\
\hline
%\hline
%
ECN~\cite{zhong2019invariance}  &$M^I$&\textit{Duke} & 0.102 & 0.302\\

DIM+GLO~\cite{dimglo} & $M^I$ & \textit{Duke} & 0.244 & 0.565 \\

%SSG~\cite{ssg}  & $M^D$ & None & 0.133 & 0.322\\

MMT~\cite{mmt} & $M^D$ & None & 0.233 & 0.501\\

%\hline
GCMT & $M^D$ & None & \textbf{0.266} & \textbf{0.579} \\
\hline

\end{tabular}}
\vspace{-3mm}
\caption{Comparison with state-of-the-art methods on \textit{MSMT}. ``S. M.'' denotes source model. ``S. D.'' denotes labeled source data.}% $^{\ddagger}$ denotes multi-scale features are used.}
\label{tab:msmt}
\end{center}
\vspace{-4mm}
\end{table}

\section{Experiment}

\subsection{Dataset}
Experiments are performed on three datasets, \textit{i.e.}, \textit{DukeMTMC-reID}~\cite{duke}, \textit{Market-1501}~\cite{market}, and \textit{MSMT17}~\cite{msmt}. %Details of these datasets are given as follows.

\textit{DukeMTMC-reID} contains 36,411 images of 1,812 identities.
% captured from 8 cameras at Duke University. 
 16,522 images of 702 identities are used for training.
 % and others are used for testing.  
 In the rest of images, 3,368 images are selected as query images and remaining 19,732 images are used as gallery images.
 
\textit{Market-1501} contains 32,668 images of 1,501 identities.
% captured from 6 cameras at Tsinghua University. 
 12,936 images of 751 identities are selected for training. 
 %and others are used for testing. 
 In the rest of images, 3,368 images are selected as query images and remaining 19,732 images are used as gallery images.

\textit{MSMT17} contains 126,441 images of 4,101 identities.
% captured from 15 cameras at Peking University. 
 32,621 images of 1,041 identities are selected for training.
 % and others are selected for testing.
 In the rest of images, 11,659 images are selected as query images and remaining 82,161 images are used as gallery images.

For short, we denote \textit{DukeMTMC-reID} as \textit{Duke}, \textit{Market-1501} as \textit{Market}, and \textit{MSMT17} as \textit{MSMT} in the rest of this paper. Following previous works~\cite{mmt}, Cumulative Matching Characteristics (CMC) and mean Average Precision (mAP) are used to evaluate the performance. Rank1, Rank5 and Rank10 accuracies in CMC are reported.

\subsection{Implementation Detail}
%Proposed GCMT method uses different source models to initialize teacher and student networks.
 Public models pre-trained on \textit{ImageNet}~\cite{imagenet}, \textit{CUHK03}~\cite{cuhk03} (14,096 images with 1,467 identities for training), \textit{Duke}, \textit{Market}, and \textit{MSMT17} are obtained following~\cite{mmt} and denoted as $M^I$, $M^C$, $M^D$, $M^{M}$, and $M^{MS}$, respectively. These pre-trained models adopt ResNet50~\cite{resnet} as backbone. 
 Input images are resized to 256$\times$128. We use random flipping, random cropping, and random erasing~\cite{zhong2017random} for data augmentation.
 We feed the same image batch to different pairs of teacher and student networks but with separately random augmentation.
 K-Means method is used for unsupervised clustering. The number of clusters is set to 500 on \textit{Duke} and \textit{Market} and 1,500 on \textit{MSMT} following~\cite{mmt,mebnet}.
 In each training batch, 16 identities are randomly selected and 4 images for each identity are selected, resulting 64 images. 
 $K$ is set as 12 in teacher graph construction.
  Loss weight $\lambda_{GCC}$ is set as 0.6. $\beta$ is set to 0.05 following~\cite{dimglo}. 
 Adam optimizer is used for training. Learning rate is initialized as 0.00035 and decayed by 0.1 after 20 epochs. Models are totally trained for 120 epochs with 400 iterations in each epoch. 
 After training, teacher networks are used to extract features for ReID and the best performance among different teacher networks is reported.
 Models are trained on a server with three GeForce RTX 2080 Ti GPUs and one Tesla V100 GPU.
% one E5-2620 CPU, 
% and 128G memory. 
% Training costs around 12 hours with one pre-trained model, and 30 hours with three pre-trained models.

 \subsection{Compared with State-of-the-art Methods}
 \label{sec:compared}
 In this section, proposed GCMT method is compared with many state-of-the-art methods. For fair comparison, GCMT uses only one pre-trained model in this section. 
 The comparison results are summarized in Table~\ref{tab:compared-others} and Table~\ref{tab:msmt}.
 It can be observed that proposed GCMT method outperforms previous methods by a clear margin.
%  which verifies the effectiveness our method. 
  It is worth noting that GCMT with $M^I$, \textit{i.e.}, in the unsupervised setting, even outperforms most recent domain adaptive methods.% except MEB-Net$^{\S}$ that uses a much deeper network, \textit{i.e.}, DenseNet121~\cite{huang2017densely}, as backbone.
% Moreover, GCMT with $M^D$ or $M^M$ approach supervised performance on \textit{Market} or \textit{Duke}, respectively.

% \textbf{Comparison on \textit{Market} and \textit{Duke}.}
 On \textit{Market} and \textit{Duke}, proposed GCMT is compared with recent state-of-the-art methods as shown in Table~\ref{tab:compared-others}. 
 It is clear that GCMT achieves the best perform compared with others. For example, GCMT achieves 77.1\% and 67.8\% in mAP accuracy on \textit{Market} and \textit{Duke}, respectively. This outperforms MMT~\cite{mmt} by 5.9\%  and 4.7\%, respectively. MEB-Net$^{\S}$~\cite{mebnet} uses DenseNet121~\cite{huang2017densely} as backbone, while GCMT still outperforms it by 1.1\% and 1.7\% in mAP accuracy on \textit{Market} and \textit{Duke}, respectively. This indicates that proposed GCMT is powerful to learn discriminative person features without annotation.
% \textbf{Comparison on \textit{Duke}.}
  
% \textbf{Comparison on \textit{MSMT}.}
 \textit{MSMT} is more challenging than \textit{Market} and \textit{Duke} due to more identities and diverse appearance.
 We compare GCMT with state-of-the-art methods as shown in Table~\ref{tab:msmt}.
 It is clear that GCMT outperforms previous methods by a clear margin. For example, GCMT achieves 54.8\% in Rank1 accuracy when using $M^M$, outperforming DIM+GLO and MMT by 5.1\% and 5.6\%, respectively.
%  Moreover, GCMT with $M^I$ as source model achieves 0.543 in Rank1 accuracy, outperforming most methods with different source model or source labeled dataset.
 Comparison on \textit{MSMT} further demonstrates the effectiveness of proposed GCMT method on large-scale person ReID task.

\begin{table}[t]

\begin{center}
\footnotesize
%\resizebox{0.9\linewidth}{!}{
\begin{tabular}{cccccc}
\hline
$\beta$ & 0.01 & 0.03 & 0.05 & 0.07 & 0.1 \\
\hline
\textit{Market} &0.721 &0.732 & \textbf{0.739} & 0.727 & 0.691\\
\hline
\textit{Duke} & 0.611&0.621 & \textbf{0.636} & {0.633} &0.601 \\
 \hline 
\end{tabular}%}
\vspace{-3mm}
\caption{Evaluation on $\beta$.}
\label{tab:a3}
\end{center}
\vspace{-4mm}
\end{table}  
 
\begin{figure}[t]
\begin{center}
\includegraphics[width=0.85\linewidth]{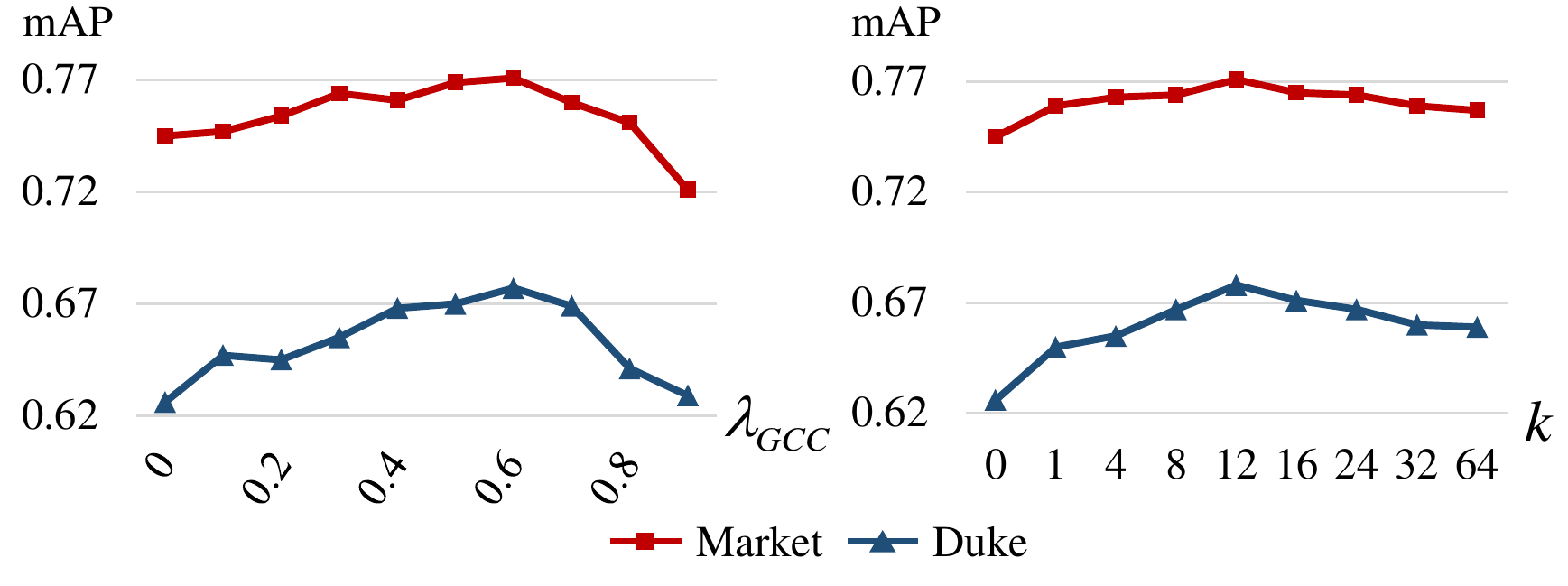}
\end{center}
\vspace{-4mm}
\caption{Parameter analysis on $\lambda_{GCC}$ and $k$ on \textit{Market} and \textit{Duke}.} 
\label{fig:lw-k}
\vspace{-3mm}
\end{figure}

\subsection{Model Analysis}
\label{sec:ma}
In this section, we first analysis the effect of $\beta$, $\lambda_{GCC}$ and $K$ respectively by varying the value of one parameter and keeping others fixed to the optimal value. Experiments show that hyper-parameters selected on one dataset can be applied to others. We then evaluate proposed GCC to show its validity. Finally, we show that GCMT could effectively boost the performance with multiple pre-trained models.

%\textbf{Analysis of $\beta$.} 
Evaluation on $
\beta$ is performed on \textit{Market} and \textit{Duke} using $M^I$ as source model, as shown in Table~\ref{tab:a3}. It is clear that setting $\beta$ to 0.05 achieves the best performance.

%\textbf{Analysis of $\lambda_{GCC}$.} 
 Evaluation on $\lambda_{GCC}$ is performed on \textit{Market} and \textit{Duke} with pre-trained models $M^D$ and $M^M$, respectively. The experimental results are shown in Fig.~\ref{fig:lw-k}. It can be observed that setting $\lambda_{GCC}$ in the range from 0.1 to 0.7 can always improves the performance and the best performance is achieve when setting $\lambda_{GCC}$ to 0.6 on both dataset.

%\textbf{Analysis of $K$.}
 Evaluation on $K$ is also performed on \textit{Market} and \textit{Duke} with pre-trained models $M^D$ and $M^M$, respectively. The evaluation results are shown in Fig.~\ref{fig:lw-k}. Setting $K=0$ denotes method without $\mathcal{L}_{GCC}$. It can be observed that setting $K$ larger than 0 always improves the performance compared with $K=0$ and the best performance is achieve when $K=12$ on both datasets. This shows the validity of using GCC in GCMT for feature learning.

% \textbf{Analysis of GCC.}
  Evaluation on proposed GCC is performed on  \textit{Market} and \textit{Duke} with pre-trained models $M^D$ and $M^M$, respectively. The comparison results are shown in Table~\ref{tab:ablation}. It can be observed that GCC improves the performance on both datasets. For example, GCC improves mAP accuracy by 2.6\% and 5.2\% on \textit{Market} and \textit{Duke}, respectively. It is also clear that GCC outperforms soft triplet loss used in~\cite{mebnet}, \textit{e.g.}, by 3.6\% in mAP on \textit{Duke}.% This shows the advantage of proposed GCC over soft triplet loss.

% \textbf{GCMT with multiple pre-trained models.}
 With multiple pre-trained models, multiple pairs of teacher and student networks are used in training. 
 The performance comparison is shown in Table~\ref{tab:multiple}. 
 It can be observed that, given multiple pre-trained models, GCMT is able to effectively boost the performance. For example, GCMT improves the mAP accuracy to 78.6\% on \textit{Market} by using two pre-trained models $M^D$ and $M^{MS}$. GCMT further boosts the mAP accuracy to 79.7\% on \textit{Market} by additionally using $M^C$.
 It is worthy noting that the improvement of using multiple source models is based on high performance by single source model, which approaches the supervised baseline as shown in Table~\ref{tab:compared-others}. Thus, the improvement is significant though it may be of small scale.
 We also implement MEB-Net~\cite{mebnet} method with three pre-trained models by provided source code. It can be observed that GCMT outperforms MEB-Net by a clear margin, \textit{e.g.}, by 6.6\% in mAP on \textit{Market}. This further shows the advantage in teacher networks cooperation of our method. 
%  Considering \textit{ImageNet} is different from person datasets, adding $M^I$ will not  offer much improvement over other pre-trained models. Experiments show that ``$M^I \! + \! M^D \! +  \! M^C \! + \! M^{MS}$'' achieves 79.8\% and 69.3\% in mAP accuracy on \textit{Market} and \textit{Duke}, respectively, which shows marginal improvements compared with the case not using $M^I$. 
 
%ablation
\begin{table}[t]
%\vspace{-5mm}
\begin{center}
\setlength{\tabcolsep}{0.2cm}
\resizebox{1.\linewidth}{!}{
\begin{tabular}{l|c|c||c|c}
\hline
%\multirow{2}{*}{Method} & \multirow{2}{*}{Reference} & \multicolumn{4}{c||}{Market-1501} & \multicolumn{4}{c}{DukeMTMC-reID}\\
\multirow{2}{*}{Method} & \multicolumn{2}{c||}{Market} & \multicolumn{2}{c}{Duke}\\

\cline{2-5}
% & \multicolumn{2}{c|}{From $M^I$} & \multicolumn{2}{c||}{From $M^D$} & \multicolumn{2}{c|}{From $M^I$} & \multicolumn{2}{c}{From $M^M$}  \\
  & mAP & Rank1& mAP &Rank1\\
\hline 
%Direct transfer & 0.289 & 0.579 & 0.280 & 0.453 \\
%\hline
w/o $\mathcal{L}_{GCC}$ &0.745 & 0.887 &0.626 & 0.781\\
Replace $\mathcal{L}_{GCC}$ with soft triplet & 0.757 & 0.891& 0.642&0.778\\
GCMT & \textbf{ 0.771} & \textbf{0.906} & \textbf{0.678}& \textbf{0.811}\\
\hline
%Supervised baseline & 0.811 & 0.930 & 0.704 & 0.848\\
%\hline
\end{tabular}}
\vspace{-3mm}
\caption{Evaluation on GCC  on \textit{Market} and \textit{Duke}.}
\label{tab:ablation}
\end{center}
\vspace{-4mm}
\end{table}

%ablation
\begin{table}[t]
%\vspace{-5mm}
\begin{center}
\setlength{\tabcolsep}{0.2cm}
\resizebox{1.\linewidth}{!}{
\begin{tabular}{l|l|c||l|c}
\hline
%\multirow{2}{*}{Method} & \multirow{2}{*}{Reference} & \multicolumn{4}{c||}{Market-1501} & \multicolumn{4}{c}{DukeMTMC-reID}\\
\multirow{2}{*}{Method} & \multicolumn{2}{c||}{Market} & \multicolumn{2}{c}{Duke}\\

\cline{2-5}
% & \multicolumn{2}{c|}{From $M^I$} & \multicolumn{2}{c||}{From $M^D$} & \multicolumn{2}{c|}{From $M^I$} & \multicolumn{2}{c}{From $M^M$}  \\
  & S. M. & mAP & S. M. &mAP \\
\hline 
GCMT & $M^D$ & 0.771 & $M^M$ &0.678 \\
GCMT & $M^{MS}$ &0.776  & $M^{MS}$ &0.681 \\
GCMT & $M^C$ & 0.752  & $M^C$ &0.637 \\
\hline
GCMT & $M^D + M^{MS}$ &  0.786 & $M^M + M^{MS}$ & 0.688\\
GCMT & $M^D+M^C + M^{MS}$ & \textbf{0.797} &  $M^M+M^C+M^{MS}$& \textbf{0.691}\\
\hline
MEB-Net$^*$ & $M^D+M^C + M^{MS}$ & 0.731 & $M^M+M^C+M^{MS}$  & 0.636\\

\hline
\end{tabular}}
\vspace{-3mm}
\caption{Performance comparison with different source models on \textit{Market}  and \textit{Duke}. $^*$ denotes our own implementation.}
\label{tab:multiple}
\end{center}
\vspace{-4mm}
\end{table}

\section{Conclusion}
This paper proposes a Graph Consistency based Mean-Teaching (GCMT) method for unsupervised domain adaptive person ReID task.
 GCMT fuses sample similarity relationships predicted by different teacher networks as supervise signal based on graphs. With fused sample similarity relationships, GCMT uses the proposed Graph Consistency Constraint (GCC) to train student networks with more sample similarity relationships involved.  
 GCMT provides an effective method for teacher networks cooperation and student networks optimization.
 Extensive experiments show that our GCMT  outperforms state-of-the-art methods by a clear margin.
% Specially, by only using the model pre-trained on \textit{ImageNet}, our method still outperforms most recent works.
 Experimental results also show that GCMT effectively improves the performance given multiple pre-trained models. 

\section*{Acknowledgments}
This work is supported in part by Peng Cheng Laboratory, in part by The National Key Research and Development Program of China under Grant No. 2018YFE0118400, in part by Natural Science Foundation of China under Grant No. U20B2052, 61936011, 61620106009, in part by Beijing Natural Science Foundation under Grant No. JQ18012.


\bibliographystyle{named}
\bibliography{ijcai21}

\end{document}